\documentclass{article}

\usepackage[preprint, nonatbib]{neurips_2022}

\usepackage[utf8]{inputenc} 
\usepackage[T1]{fontenc}    
\usepackage{hyperref}       
\usepackage{url}            
\usepackage{booktabs}       
\usepackage{amsfonts}       
\usepackage{nicefrac}       
\usepackage{microtype}      
\usepackage{xcolor}         

\usepackage{graphicx}
\usepackage{amsmath,amssymb}
\usepackage{wrapfig}
\usepackage[font=small]{caption}

\usepackage{enumitem}
\setlist{leftmargin=4mm, itemsep=0.1pt}

\title{A generative recommender system with GMM prior\\
for cancer drug generation and sensitivity prediction}

\author{%
  Krzysztof Koras \\
  Faculty of Mathematics, Informatics and Mechanics\\
  University of Warsaw\\
  \And
  Marcin Możejko \\
  Faculty of Mathematics, Informatics and Mechanics\\
  University of Warsaw\\
  \And
  Paulina Szymczak \\
  Faculty of Mathematics, Informatics and Mechanics\\
  University of Warsaw\\
  \And
  Eike Staub \\
  Merck KGaA, Translational Medicine\\
  Department of Bioinformatics\\
  \And
  Ewa Szczurek\thanks{Corresponding author} \\
  Faculty of Mathematics, Informatics and Mechanics\\
  University of Warsaw\\
  \texttt{szczurek@mimuw.edu.pl} \\
}

\begin{document}

\maketitle

\begin{abstract}
Recent emergence of high-throughput drug screening assays sparkled an intensive development of machine learning methods, including models for prediction of sensitivity of cancer cell lines to anti-cancer drugs, as well as methods for generation of potential drug candidates. 
However, a concept of generation of compounds with specific properties and simultaneous modeling of their efficacy against cancer cell lines has not been comprehensively explored. To address this need, we present VADEERS, a Variational Autoencoder-based Drug Efficacy Estimation Recommender System. The generation of compounds is performed by a novel variational autoencoder with a semi-supervised Gaussian Mixture Model (GMM) prior. The prior defines a clustering in the latent space, where the clusters are associated with specific drug properties. In addition, VADEERS is equipped with a cell line autoencoder and a sensitivity prediction network. The model combines data for SMILES string representations of anti-cancer drugs, their inhibition profiles against a panel of protein kinases, cell lines’ biological features and measurements of the sensitivity of the cell lines to the drugs. The evaluated variants of VADEERS achieve a high $r=0.87$ Pearson correlation between true and predicted drug sensitivity estimates. We train the GMM prior in such a way that the clusters in the latent space correspond to a pre-computed clustering of the drugs by their inhibitory profiles. We show that the learned latent representations and new generated data points accurately reflect the given clustering. In summary, VADEERS offers a comprehensive model of drugs’ and cell lines’ properties and relationships between them, as well as a guided generation of novel compounds.

\end{abstract}

\section{Introduction}
Kinase inhibitors are a class of anticancer drugs that target specific mutated kinases and disregulated biological processes in tumor cells~\cite{comprehensive_review_of_prot_kinases}. As such, they constitute flagship examples of personalized cancer treatments~\cite{prot_kinases_review_bmc, prot_kinase_2020_survey}. The set of  kinase inhibitors is deeply investigated experimentally. 
First, they are commonly characterized by their {\textit{inhibition profiles}}, measuring their strength of inhibition of a vector of kinases~\cite{kinase_profiling1, kinase_profiling2}. Second, large-scale experiments were performed, measuring the sensitivity of cancer cell lines to these and other cancer compounds~\cite{gdsc_database1, ccle_database, ctrp_database}. Third, the molecular features of the cancer cell lines, such as gene mutations 
and gene expression were measured~\cite{gdsc_database1, ccle_database, gdsc_database2}. Despite their limitations, cancer cell lines commonly act as laboratory proxies for patients' tumors and it is known that their molecular features are key determinants of their response to anticancer drugs~\cite{ccle_database, relevance_ccl}. Finally, kinase inhibitors can be grouped into different clusters, for example by their target pathways,
or kinase inhibitory activity. Specifically, it has been observed that they differ by the numbers of so called off-targets, i.e. unintentionally inhibited kinases, which is reflected in their inhibitory profiles. While a number of  kinase inhibitor drugs is already successfully applied in the clinic, there is a pressing need for novel drug discovery, due to the mechanism of resistance to existing drugs and the large variety of mutations that could be targeted in individual patients' tumors. 

Unfortunately, the current pre-clinical process of proposing novel compounds proves inefficient, as the proposed drugs fail further stages of clinical trials, yielding the entire process of novel drug discovery a daunting, time and money consuming task~\cite{ai_drug_disc1, ai_drug_disc2, ai_drug_disc3}. Deep generative models transform the field of molecule discovery,  providing promising synthetic molecules such as proteins or drugs with desired chemical properties~\cite{gen_models_drugs1, gen_models_drugs2, gen_models_drugs3, gen_models_drugs4, gen_models_drugs5, gen_models_drugs6, gen_models_drugs7}. However, 
these approaches are not directly 
applicable to kinase inhibitors. First, they require large amounts of compounds for training, while the number of known kinase inhibitors is scarce. Second, they do not account for the  molecular features of tumors that the drugs are supposed to act on. Specifically, drug sensitivity is a function of both compound's and tumor's features, and it is the relationship between these two features sets that determines the treatment outcome. Although multiple machine learning models were proposed for the prediction of the sensitivity of cancer cell lines to the drugs~\cite{azuaje_review, drp_review_nature_2020, overview_of_ML_for_monotherapy, jang, nci_dream}, including recommender system-based approaches~\cite{brandao_et_al_2021, deers, chayaporn_et_al_2018}, these methods lack a generative ability. Therefore, a new class of generative models for kinase inhibitor discovery and simultaneous sensitivity prediction is needed, which would restrict the vast space of potential generative model hypotheses by accounting for and drawing from the wealth of additional experimental data such as kinase inhibitor profiles and their clustering, cell line sensitivity screens, as well as the molecular profiling of the cell lines.

In this work, we propose a novel variational autoencoder (VAE) model for generation of specific types of kinase inhibitors, guided by the clustering of their inhibitory profiles within the GMM prior.
The model infers 
representations of reduced dimensionality of the drugs' SMILES~\cite{smiles} (a string-based representation of molecule's chemical structure), and leverages drug sensitivity screening data, as well as molecular features of the cancer cell lines. The proposed model offers the following key functionalities:
\begin{itemize}
    \item \textbf{clustering of the drugs in the latent space and generation of novel drugs from specific clusters}, here having specific types of inhibitory profiles,    
    \item \textbf{inference of latent representations} of the drugs' SMILES and the molecular features of the cell lines,
    \item \textbf{prediction of sensitivity of the cancer cell lines to the drugs}.
\end{itemize}

On the most general level, the proposed model can be thought of as an extension of a recommender system with side information~\cite{rs_w_side_info1, rs_w_side_info2, rs_w_side_info3, rs_w_side_info4, rs_w_side_info5} 
with a generative model. 
In a recommender system with side information, objects and users are characterized by vectors of their specific features (the side information), and users assign scores to objects, yielding an object per user matrix. The task of a recommender system is to predict the scores given the side information for the objects and users. 
Our contributed extension is twofold. First, we assume the objects can be grouped into clusters, 
and we generate new objects with features that are characteristic of their corresponding clusters. Second, we aim to predict the scores  for these synthetic objects and for existing users. In our particular application, in the generative recommender system the objects correspond to drugs from the family of kinase inhibitors, users to cancer cell lines, while the scores correspond to the sensitivity of the cell lines to the drugs, i.e. the ability of the drugs to kill cancer cells. Hence the name of the model, i.e, Variational Autoencoder-based Drug Efficacy Estimation Recommender System (VADEERS).

\section{Related work}
\subsection{Deep generative models for anticancer drug discovery}
The majority of existing methods for compound generation focus on generating compounds with specified chemical properties, without taking into account the broader biological context, e.g. efficacy of compounds against cancer cell lines or other cancer models. Recently, Born \textit{et al.}~\cite{paccmann_rl} proposed a model aiming at generating compounds which target specific gene expression profiles via a hybrid variational autoencoder acting as compounds generator. In the proposed reinforcement learning paradigm, the compound generator serves as an agent, while the output of a drug sensitivity prediction model serves as a reward function, which allows to train the agent to produce more and more effective compound against a given gene expression profile. Joo \textit{et al.}~\cite{joo_et_al} proposed a conditional VAE, in which generation of new molecular fingerprints is conditioned on drug sensitivity. Related problem was also approached without resolving to probabilistic generative models by efficient Monte Carlo tree search~\cite{monte_carlo_search}. VAEs have also appeared in the context of drug sensitivity modeling focusing more on the prediction aspect~\cite{Dong_et_al}, or applied to cancer models, rather than compounds~\cite{dr_vae, Jia_et_al_2021}.

\subsection{Recommender systems with generative components}

Recommender systems were traditionally applied in the field of e-commerce, where user decisions are recorded online. Generative adversarial network-based models proved useful in dealing with the lack of explicitly negative samples in such applications~\cite{generative_rs1, generative_rs2, generative_rs3, generative_rs4}. GANs were also used to imitate user behavior dynamics in reinforcement learning-based recommender systems to better approximate the reward function and simulate the environment in this setting~\cite{generative_rs5, generative_rs6}. We are not aware of recommender systems equipped with VAE that accounts of a given clustering of the objects. 

\subsection{Variational Autoencoders}

Variational autoencoders \cite{vae_original} can be viewed as neural network-based probabilistic graphical models designed to learn complexed probability distributions. They are latent variable models, where sampling from a prior distribution $p(\vec{z})$ in the latent space $Z$ is followed by a sampling from $p_{\theta}(\vec{x}|\vec{z})$ in the data space $X$. The probability distribution $p_{\theta}(\vec{x}|\vec{z})$ is modeled using a neural network parametrized by $\theta$ called a decoder. Since usually the log marginal likelihood $p(\vec{x})$ is intractable, its following evidence lower bound (ELBO) is maximized: 
\begin{equation}   \label{eq:vae:elbo}
    \mathcal{L}^{ELBO}_{\phi, \theta}(X) = \mathbb{E}_{q_\phi(\vec{z} \mid \vec{x})} \big[ \ln p_\theta(\vec{x} \mid \vec{z}) \big] - D_{KL}(q_\phi (\vec{z} | \vec{x}) || p(\vec{z})),
\end{equation}
where $q_\phi (\vec{z} | \vec{x})$ is the variational approximation of a posterior distribution $p(\vec{z}|\vec{x})$, and $D_{KL}$ stands for Kullback–Leibler divergence. $q_\phi (\vec{z} | \vec{x})$ is usually modeled as a Gaussian distribution $\mathcal{N}\left(\mu(\vec{x}), diag(\sigma(\vec{x}))\right)$ where $\mu(\vec{x})$ and $\sigma(\vec{x})$ are outputs from a neural network parametrized by $\phi$, called encoder. The first term in Eq.~(\ref{eq:vae:elbo}) is often referred to as the reconstruction term, while the second as the regularization term, 
as it forces the posterior towards the prior. In case when both $q_\phi (\vec{z} | \vec{x})$ and $p(\vec{z})$, are Gaussian distributions, the Kullback-Leibler divergence has an easy to optimize analytical form~\cite{vae_original}, which in turn enables an efficient optimization of Eq.~(\ref{eq:vae:elbo}). In a more general case, Eq.~(\ref{eq:vae:elbo}) can be further decomposed into \cite{vae_vamp}:
\begin{equation}   \label{eq:vae:elbo_three_terms}
    \mathcal{L}^{ELBO}_{\phi, \theta} = \mathbb{E}_{q_\phi(\vec{z} | \vec{x})} \big[ \ln p_\theta(\vec{x} | \vec{z})  + \ln p(\vec{z}) - \ln q_\phi (\vec{z} | \vec{x})  \big].
\end{equation}
Since $\ln q_\phi(\vec{z} | \vec{x})$ is under the expectation over $q_\phi(\vec{z} | \vec{x})$, this equals:
\begin{equation}   \label{eq:vae:elbo_entropy}
    \mathcal{L}^{ELBO}_{\phi, \theta} = \mathbb{E}_{q_\phi(\vec{z} | \vec{x})} \big[ \ln p_\theta(\vec{x} | \vec{z}) + \ln p(\vec{z}) \big] + \mathbb{H} \big[ q_{\phi}(\vec{z}|\vec{x}) \big],
\end{equation}
where $\mathbb{H} \big[ q_{\phi}(z|x) \big]$ denotes the entropy of the posterior. Typically, the expected value in Eq.~(\ref{eq:vae:elbo_three_terms}) or (\ref{eq:vae:elbo_entropy}) is approximated by sampling $L$ point(s) from $q_\phi(\vec{z} | \vec{x})$~\cite{vae_vamp}. In VAEs, the goal is to maximize the ELBO, while when training neural networks in general commonly the goal is to minimize a cost function. Therefore, with $L = 1$, the loss function 
takes the form of
\begin{equation}   \label{eq:vae:loss_f_entropy}
    \mathcal{L}_{\phi, \theta} (\vec{x}^{(i)}) = - \ln{ p_{\theta}(\vec{x^{(i)}}|\vec{z}^{(i)})} - \ln{p(\vec{z}^{(i)})} - \mathbb{H} \big[ q_{\phi}(\vec{z}|\vec{x}^{(i)}) \big],
\end{equation}
where $\vec{x}^{(i)}$ indicates the $i$-th data point and $z^{(i)}$ is a sample from a $q(\vec{z}|\vec{x}^{(i)})$. In case when a variational posterior approximation is $\mathcal{N}\left(\mu(\vec{x}), diag(\sigma(\vec{x}))\right)$, the entropy has an easy to optimize analytical form. In such a case, optimization of Eq.~(\ref{eq:vae:loss_f_entropy}) is feasible for $\ln{p(\vec{z}^{(i)})}$ that are easy to evaluate.

\subsection{GMM VAEs}

In VAE, different choices of prior distributions $p(\vec{z})$ impose different trade-offs between the optimization simplicity and the complexity of modeled distributions. E.g., the classical choice of the prior to be the standard normal distribution, i.e.  $p(\vec{z}) \sim \mathcal{N} (\textrm{\textbf{0}}, \textrm{\textbf{I}})$ carries the simplicity of optimization of $D_{KL}(q_\phi (\vec{z} | \vec{x}) || p(\vec{z}))$, at the cost of not imposing any particular structure on the latent space, nor utilizing any prior knowledge regarding the data. Another popular choice of a prior $p(\vec{z})$ is a GMM \cite{vde, multi-facet, Guo2020VariationalAW, DUCWGMVA, DEFMVAE}. In this model, for a given point $\vec{z}$ there is a categorical hidden variable $C \sim Cat\left(\pi_1, \dots \pi_K\right)$,  defining the component of the mixture for that point.  The conditional probability of $\vec{z}$ given the component $C=k$, is then defined by a Gaussian distribution $\mathcal{N}\left(\mu_k, \Sigma_k\right)$, for $k = 1, \dots, K$.
Therefore, in a GMM VAE, the
prior for $z$ in Eq.~(\ref{eq:vae:elbo_entropy}) is obtained by marginalizing over the values of $C$:
\begin{equation} \label{eq:gmm_superposition}
    p(\vec{z}) = \sum_{k = 1}^{K} \pi_k \mathcal{N} (\vec{z} | \vec{\mu}_k, \Sigma_k).
\end{equation}
The closed, analytical form of Eq.~(\ref{eq:gmm_superposition}) makes computation of the $\ln p(\vec{z})$ term in Eq.~(\ref{eq:vae:elbo_entropy}) tractable and enables an efficient optimization of ELBO. 
Such a GMM prior naturally corresponds to a clustering, where the points from the same component $k$ come from the same Gaussian distribution and thus group together in the latent space. 

\begin{figure}[h!]
  \centering
  \includegraphics[width=0.80\linewidth]{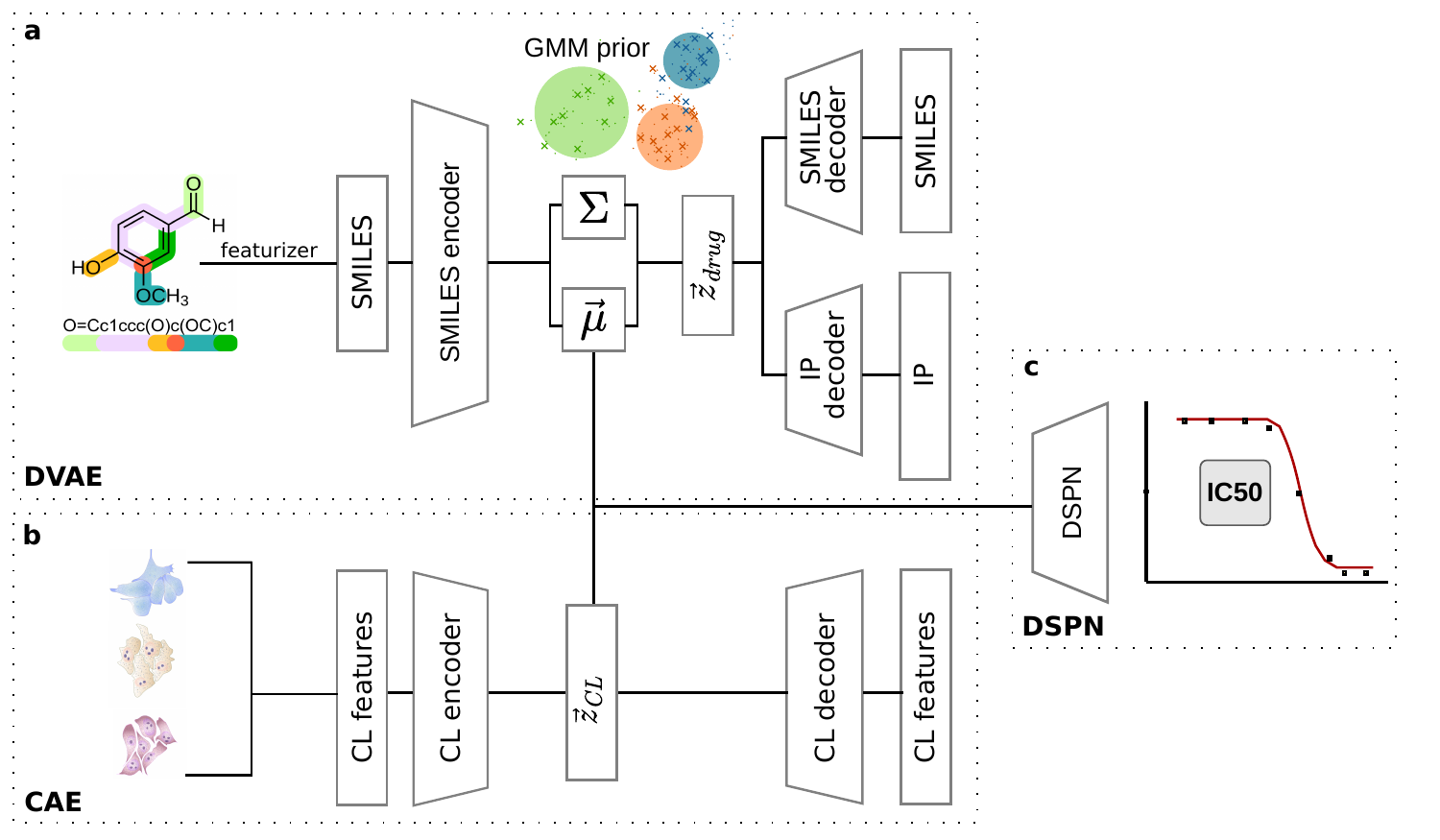}
  \caption{Model's overview. (\textbf{a}) DVAE module. (\textbf{b}) CAE module. (\textbf{c}) DSPN module. DSPN takes concatenation of DVAE's encoder output, i.e. mean vector, and CAE's latent vector as input.}
  \label{fig:overview}
\end{figure}

\section{Methods}

\subsection{Generative recommender system overview}  \label{subsection:model_overview}
The proposed VADEERS model is a neural network consisting of three major modules: drug variational autoencoder (DVAE) (Fig. \ref{fig:overview}a), cell lines autoencoder (CAE) (Fig. \ref{fig:overview}b), and drug sensitivity prediction network (DSPN) (Fig. \ref{fig:overview}c). The whole model has two inputs, one to DVAE and another to CAE, and four outputs in total. 
The input to DVAE is a vector representation of a drug's chemical formula, expressed as its SMILES string. 
DVAE consists of an encoder, which projects the data into a lower-dimensional latent space, and two decoders: the first one outputs the input's reconstruction, while the second predicts the drug's inhibition profile (IP), i.e. the vector of inhibition strengths across a panel of protein kinases.  We assume that the drugs can be grouped into clusters by some guiding data that specifies the grouping. Here, the guiding data used are the inhibitory profiles, i.e., drugs with similar inhibitory profiles 
form distinct clusters (see Section~\ref{subsection:dataset} and Supplementary Methods). CAE is an autoencoder taking vector representation of a cell line's biological features as input to an encoder, and returning its reconstruction from a decoder. 

The use of DVAE and CAE allows to find lower-dimensional, informative data representations of drugs and cell lines, respectively. During the forward pass, the latent representations of a given cell line and a given drug are extracted, concatenated and passed as an input to DSPN, which predicts the numerical value indicating the sensitivity of that cell line to that drug. Here, this value is represented by log half maximal inhibitory concentration (IC50)~\cite{ic50_ref}, defined as a drug concentration needed to reduce cell viability by 50\%.

\subsection{Extension of a GMM VAE model to unknown guiding data and unknown components}
We propose an  extension to the classical GMM VAE by allowing a semi-supervision of the GMM prior. Specifically, we consider that the categorical variables defining the mixture components for each point in the latent space are partially observed for some of the training data.  
More formally, the training input data $\mathcal{D}$ with $N$  samples  can be divided into two disjoint sets $\mathcal{D}_{o}$ and $\mathcal{D}_{h}$, with $\mathcal{D} = \mathcal{D}_{o} \cup \mathcal{D}_{h}$. The set $\mathcal{D}_{o} = \left\{x^{(1)}, \ldots, x^{(n)} \right\}$, where  $0\leq n\leq N$,
is a set of samples $x^{(i)}$ for which the component variable $C^{(i)}$ is observed and $C^{(i)}=k^{(i)}$, where $k^{(i)} \in \{1, \dots, G\}$, for $G\leq K$. 
We further refer to these observed component values as the {\textit{guiding labels}}. 
Note that with $G < K$ some of the assumed components will not appear as a guiding label for any training sample. Such components correspond to additional clusters of samples that truly exist but are never observed. By allowing additional components in the latent space, we still are able to model these clusters in the latent space. In contrast to $\mathcal{D}_{o}$, the set $\mathcal{D}_{h} = \left\{ x^{(n+1)}, \dots, x^{( N)} \right \}$, is a set of samples, for which the components are hidden. In such a setting, the latent prior $p(\vec{z})$ in Eq.~(\ref{eq:gmm_superposition}) is replaced by
\begin{equation}   \label{eq:gmm_prior}
    p(\vec{z}) = 
    \begin{cases}
        \mathcal{N}(\vec{z}|{\mu}_{k^{\ast}}, \Sigma_{k^{\ast}}) & \text{if $\vec{z}$ is a sample for input  $x^{(i)} \in \mathcal{D}_{o}$ and $C^{(i)} = k^{*}$}  \\
        \sum^K_{k=1}\pi_k\mathcal{N}(\vec{z} \mid {\vec{\mu}}_k,\Sigma_k) & {\text{otherwise}},
    \end{cases}
\end{equation}
where $k^\ast$ indicates the particular Gaussian component (cluster), $K$ is the total number of components, and $\pi_k$ is the weight of  component $k$. Note that the prior defined by Eq.~(\ref{eq:gmm_prior}) is a generalization of the classical GMM prior defined by Eq.~(\ref{eq:gmm_superposition}). Indeed, Eq.~(\ref{eq:gmm_prior})  reduces to Eq.~(\ref{eq:gmm_superposition}) in the case when  $\mathcal{D}_{o} = \emptyset$.

The guiding labels can be defined by an independent, given clustering of the samples in some external data space. In such a case, the external data is referred to as $\textit{guiding data}$.

The described model with the GMM prior is used to implement the DVAE in VADEERS (Fig. ~\ref{fig:overview}a). Here,  the input samples $x^{(i)}$ correspond to SMILES, the guiding data is defined by the inhibitory profiles of the drugs, and the guiding labels by the clusters of these inhibitory profiles. 
In this way,  we transfer a clustering of inhibition profiles to the latent space, as the latent representation of drugs sharing $k^{\ast}$ are made to follow the same Gaussian distribution in $Z$.

Importantly, the parameters of GMM, i.e. $(\pi_k, \vec{\mu}_k, \Sigma_k)$ for $k$ in $\{1 , \ldots , K \}$, can be learned via gradient descent together with the remaining parameters of VADEERS, yielding a complete model of the data, including the GMM prior. The use of GMM as a prior then allows to sample $\vec{z}$ from a particular of component of $p(\vec{z})$ e.g. corresponding to a particular guiding label.
At the same time, for the drug $\vec{x}$ for which the guiding label is unknown, the cluster assignment is obtained using a posterior inference over $C$ based on the values sampled from $q(\vec{z}|\vec{x})$. This enables the inference of the guiding label (i.e., the category of inhibition profiles) for every drug.

\subsection{VADEERS model's cost function}

Recall from section \ref{subsection:model_overview} that DVAE part of VADEERS has two decoders corresponding to reconstructed compounds' input and compounds' inhibition profiles (IP). Combining \ref{eq:vae:loss_f_entropy} with the fact that decoders' outputs are continous, we define the DVAE's loss function for a single compound as:
\begin{align}   \label{eq:DVAE_losss}
    \mathcal{L}_{DVAE}({\vec{x}_S}^{(i)}, {\vec{x}_S}^{(i)^{'}}, {\vec{x}_I}^{(i)}, {\vec{x}_I}^{(i)^{'}}, {\vec{z}}^{(i)})
    &= r_{S} \cdot \textrm{MSE}({\vec{x}_S}^{(i)}, {\vec{x}_S}^{(i)^{'}})  
    + r_{I} \cdot \textrm{MSE}({\vec{x}_I}^{(i)}, {\vec{x}_I}^{(i)^{'}})\\ 
    &- r_{P} \cdot \ln p_{\lambda}({\vec{z}}^{(i)}) 
    - r_{E} \cdot \mathbb{H} \big[ q_{\phi}(\vec{z}|{\vec{x}_S}^{(i)}) \big],
\end{align}
where ${\vec{x}_S}^{(i)}$ and ${\vec{x}_S}^{(i)^{'}}$ are the $i$-th compound's SMILES representation and it's reconstruction, respectively, ${\vec{x}_I}^{(i)}$ and ${\vec{x}_I}^{(i)^{'}}$ are true and predicted inhibition profiles, respectively, ${\vec{z}}^{(i)}$ is a $\vec{z}$ sample corresponding to the $i$-th compound, $r_S$ is the positive real-valued weight corresponding to the compounds' input reconstruction error, $\textrm{MSE}$ denotes mean squared error, $r_I$ is the weight of the IP prediction error, $r_{P}$ is the weight corresponding to the prior likelihood, $r_E$ is the weight of encoder's entropy, and the last term corresponds to the entropy of latent variables returned by the encoder. 

In the case of CAE, the loss function is given simply by the reconstruction error between the cell lines' input and its reconstruction:
\begin{equation}
    \mathcal{L}_{CAE}({\vec{x}_B}^{(j)}, {\vec{x}_B}^{(j)^{'}}) = \textrm{MSE}({\vec{x}_B}^{(j)}, {\vec{x}_B}^{(j)^{'}}),
\end{equation}
where ${\vec{x}_B}^{(i)}$ and ${\vec{x}_B}^{(i)^{'}}$ are cell line's input features and their reconstruction, respectively. Finally, the loss corresponding to DSPN is the error between continuous true and predicted IC50 values defined for compound-cell line pair:
\begin{equation}
    \mathcal{L}_{DSPN}(y^{(i, j)}, \hat{y}^{(i, j)}) = MSE(y^{(i, j)}, \hat{y}^{(i, j)}),
\end{equation}
where $y^{(i, j)}$ and $\hat{y}^{(i, j)}$ is a true and predicted IC50 corresponding to $i$th compound and $j$th cell line, respectively.
The loss for the whole model is the weighted sum of above expressions:
\begin{equation}   \label{eq:whole_model_loss}
    \mathcal{L}_{Model}(\cdot) = \mathcal{L}_{DVAE}(\cdot) + r_{CAE} \cdot \mathcal{L}_{CAE}(\cdot) + r_{DSPN} \cdot \mathcal{L}_{DSPN}(\cdot),
\end{equation}
where $r_{CAE}$ and $r_{DSPN}$ are weights corresponding to CAE and DSPN errors, respectively (arguments are replaced by $\cdot$ for simplicity). The formulation with the vector $r$ of weights allows to change model's emphasis by controlling these hyperparameters. 

\subsection{Dataset}   \label{subsection:dataset}
The analyzed dataset $\mathcal{D} = \{ \textrm{\textbf{X}}_S, \textrm{\textbf{X}}_I, \textrm{\textbf{X}}_B, \textrm{\textbf{Y}}_R, \vec{y}_G \}$ consists of five parts, where $\textrm{\textbf{X}}_S \in \mathbb{R}^{304 \times 300}$ denotes drugs' SMILES vector representations,
$\textrm{\textbf{X}}_I \in \mathbb{R}^{117 \times 294}$ denotes drugs' inihibition profiles across a panel of protein kinases, 
$\textrm{\textbf{X}}_B \in \mathbb{R}^{922 \times 241}$ denotes a matrix of cell lines biological features, 
$\textrm{\textbf{Y}}_R \in \mathbb{R}^{922 \times 304}$ denotes a matrix with drug response indicators for a given cell line $c$ and drug $d$, 
and $\vec{y}_G \in \mathbb{R}^{117}$ denotes a vector of guiding labels for a subset of considered drugs (see below). 

The primary source of drug sensitivity data for cell lines was the Genomics of Drug Sensitivity in Cancer (GDSC) database~\cite{gdsc_database1, gdsc_database2}. The set of 304 compounds in $\textrm{\textbf{X}}_S$ extracted from GDSC database were represented by their chemical structure indicated by corresponding SMILES strings. In order to convert SMILES strings into numerical vector representations, we used the pre-trained Mol2vec model~\cite{mol2vec} treating it as SMILEs featurizer which produces $300$-dimensonial vectors of continous values. These representations served as an input to DVAE. 

Another considered features of compounds were their inhibition profiles, i.e., their binding strengths across a panel of 294 protein kinases. Such inhibitory profiles were available and extracted for 117 compounds from the HMS LINCS KINOMEscan data resource~\cite{KINOMEscan_data_website}. The value for a given compound-kinase pair represents a percent of control, where a value of 100\% means no inhibition of kinase binding to its ligand in the presence of the compound, and where low value means a strong inhibition~\cite{KINOMEscan_protocol_paper, KINOMEscan_protocol_website}. 

Data to characterize the 922 cell lines were downloaded from the GDSC. For the molecular features of the cell lines, we considered only the genes coding for kinases present in the KINOMEscan dataset, as well as subset of putative gene targets of considered compounds. This resulted in a set of 202 genes, for which mRNA expression levels (202 features) and binary mutation calls (21 features) were extracted for all cell lines. Furthermore, the dummy-encoded tissue type was added, producing additional 18 binary features, yielding the final set of 241 biological features for 922 cell lines.

For the drug response indicators in $\textrm{\textbf{Y}}_R$ we used the log half maximal inhibitory concentration (IC50)~\cite{ic50_ref} values from GDSC. For a given compound-cell line pair, IC50 is defined as a drug concentration needed to reduce cell viability by 50\%. Note that some values in $\textrm{\textbf{Y}}_R$ were missing since not every cell line is screened against every available compound. 

In principle, guiding labels in $\vec{y}_G$ could be any discrete class assignments for compounds. In this case, we utilized inhibition profiles from $\textrm{\textbf{X}}_I$ to assign compounds to their functional categories. To this end, compounds were clustered according to their inhibition profiles using K-means, with the number of clusters set to $G = 3$. Cluster assignments resulting from this approach were then used as the guiding labels for the 117 compounds with known inhibition profiles. See Supplementary Methods for more details on the dataset.

\subsection{Experimental setup, VADEERS model architecture, training and implementation}
The validation and test sets were constructed by randomly selecting two sets of 100 unique cell lines each. We then extracted the compound-cell line datapoints containing selected cell lines and used them to construct validation and test sets, while the rest of the pairs corresponding to the remaining 722 unique cell lines constituted the training set.

The hyperparameters of the model were empirically determined using the validation set. The encoders for both DVAE and CAE were fully-connected forward networks with two hidden layers with 128 and 64 neurons, respectively. All of the decoders followed a similar architecture, but with 64 neurons in a first hidden layer and 128 in a second. The latent space dimensionality in both DVAE and CAE was set to 10. 

The DSPN was a fully-connected network with three hidden layers with 512, 256 and 128 neurons, respectively, and an output layer outputting an IC50 prediction. Dropout with $p = 0.5$ was applied at the first and second layer. ReLu activation function was used throughout the whole model.

Model training was performed on 200 epochs using the Adam optimizer~\cite{adam_optimizer} with a learning rate of 0.005 and batch size of 128. The whole model was trained together for the first 150 epochs, after which, DVAE and CAE were frozen and DSPN alone was trained for another 50 epochs with a newly set learning rate of 0.001, decreasing by a factor of 0.1 with every 10 epochs. In addition, every 1000 training steps there was a break devoted to only training DVAE part. During each break, DVAE was trained for 100 epochs using only compounds with known inhibition profiles, with the batch size of 8. For the purpose of experiments, the $r$ loss function weights were all set to 1. Both the number of unique guiding labels and components in GMM prior were set to 3, i.e, $G = K = 3$. 

The neural networks related code was implemented using Python 3.8.8, PyTorch 1.10.0 and PyTorch Lightning 1.5.0. K-means clustering for the guiding data was implemented using scikit-learn 1.0.1~\cite{scikit-learn}.

\section{Results}
We evaluated three versions of the proposed model, differing by the way the DVAE module was implemented: i) a classical VAE with the standard normal prior ("Vanilla VAE"), ii), the DVAE as described in Section 3.2. (with GMM prior and loss function given by Eq. (\ref{eq:DVAE_losss}) and (\ref{eq:whole_model_loss}), however, only weights $\pi_k$'s and components' means $\mu_k$'s were the trainable parameters of the GMM prior (Eq. \ref{eq:gmm_prior}), while components' covariance matrices $\Sigma_k$'s set to be identity matrices ("GMM VAE constrained"), iii) the DVAE was as described in Section 3.2., in its least constrained version, where all parameters of the GMM, including $\Sigma_k$'s, were trainable ("GMM VAE unconstrained"). 

\begin{table}[h!] \label{tab:numerical_results}
  \caption{IC50 and IP prediction performance for VADEERS with different versions of the DVAE module.}
  \label{sample-table}
  \centering
  \begin{tabular}{llll}
    \toprule
    DVAE version & IC50 RMSE     & IC50 Pearson & IP RMSE\\
    \midrule
    Vanilla VAE                 & $1.33 \pm 0.022$  & $0.87 \pm 0.006$  & $1.13 \pm 0.109$ \\
    GMM VAE constrained    & $1.33 \pm 0.023$ &  $0.87 \pm 0.006$  & $1.09 \pm 0.062$ \\
    GMM VAE unconstrained     & $1.34 \pm 0.012$       & $0.87 \pm 0.004$ & $1.04 \pm 0.030$\\
    \bottomrule
  \end{tabular}
\end{table}

\subsection{Predictive performance}
The predictive performance of the IC50 estimation was assessed by calculating the root mean squared error (RMSE) and Pearson correlation between the true and predicted IC50 values. In addition, we computed the RMSE between the true and predicted inhibition profiles, corresponding to the second decoder in DVAE (Table~\ref{sample-table}). This procedure was repeated five times with different random data splits (see Methods).

Despite the large differences in the complexity of the prior distribution, the three model versions perform almost equally well in terms of IC50 prediction. This suggest that models achieved the limit of predictive performance for this particular dataset. 
Although the optimization of the IC50 prediction was not the main goal of this study, the low RMSE and high correlation indicate that VADEERS correctly captures drug and cell line features and reliably predicts the sensitivity of unseen cancer cell lines to kinase inhibitor drugs. 

In contrast to IC50, the three model versions do differ in terms of the ability to reconstruct inhibition profiles (Table 1), measured by RMSE between true and reconstructed IPs. The best result in that regard is achieved by the GMM VAE unconstrained model ($\textrm{RMSE} = 1.04$). 
Such a result is expected, as lower constraints on the latent space representations make it easier for the model to optimize this metric. However, note that in the described setup the IP RMSE metric was computed for the training data, therefore it should rather be interpreted as model's ability to converge w.r.t. this particular metric than model's predictive performance. Still, this result suggests that the nature of the latent space is important with regard to the decoding, in the sense that this task is not entirely dependent on the decoder alone and is improved by imposing a latent space's structure. 

\subsection{Latent space structure}

\begin{wrapfigure}{r}{0.66\textwidth}

  \centering
  \includegraphics[width=0.9\linewidth]{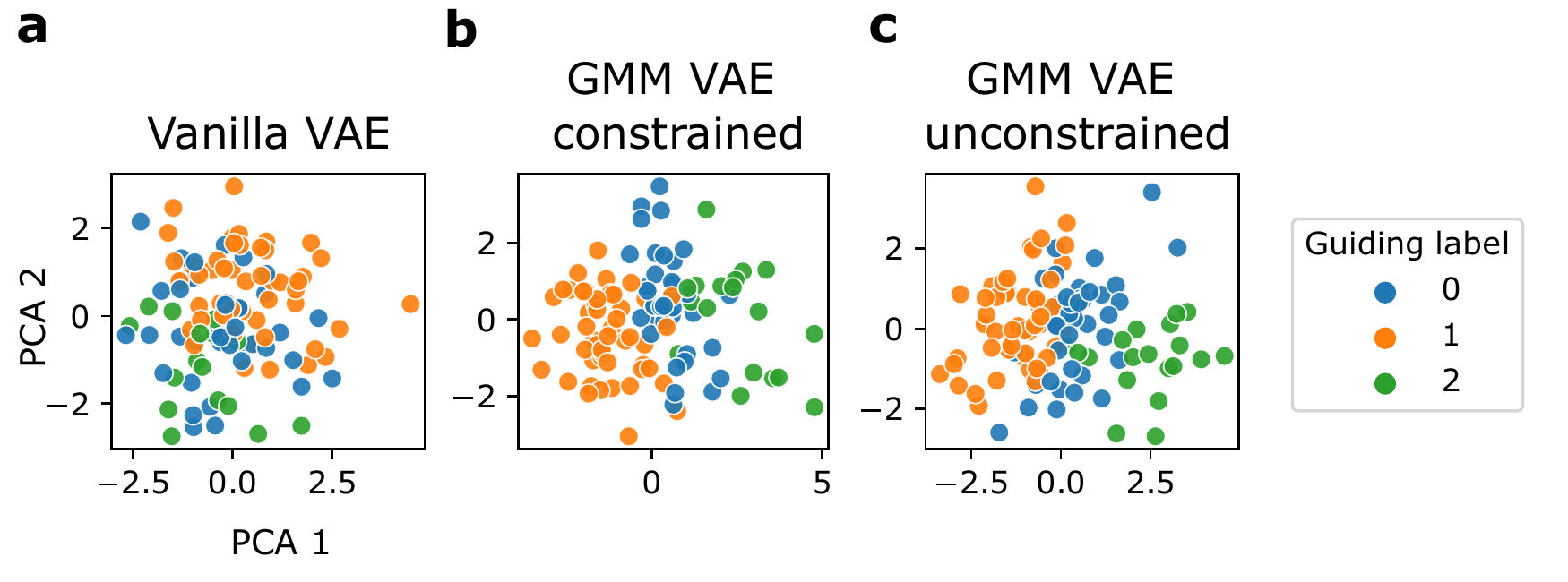}
  \caption{Latent spaces of the three VADEERS model versions, differing by the DVAE subnetwork. For each model, the latent representations of the 117 compounds with guiding labels (see Section~\ref{subsection:dataset}) are obtained by passing their SMILES representations to the model's DVAE's encoder. The encoder's output for a given model is then plotted in 2D using principal components analysis (PCA) and colored with the corresponding guiding labels.
  Results for the first of five random data splits.}
  \label{fig:latent_spaces}
\end{wrapfigure}

\begin{figure}[!h]  
  \centering
  \includegraphics[width=0.75\linewidth]{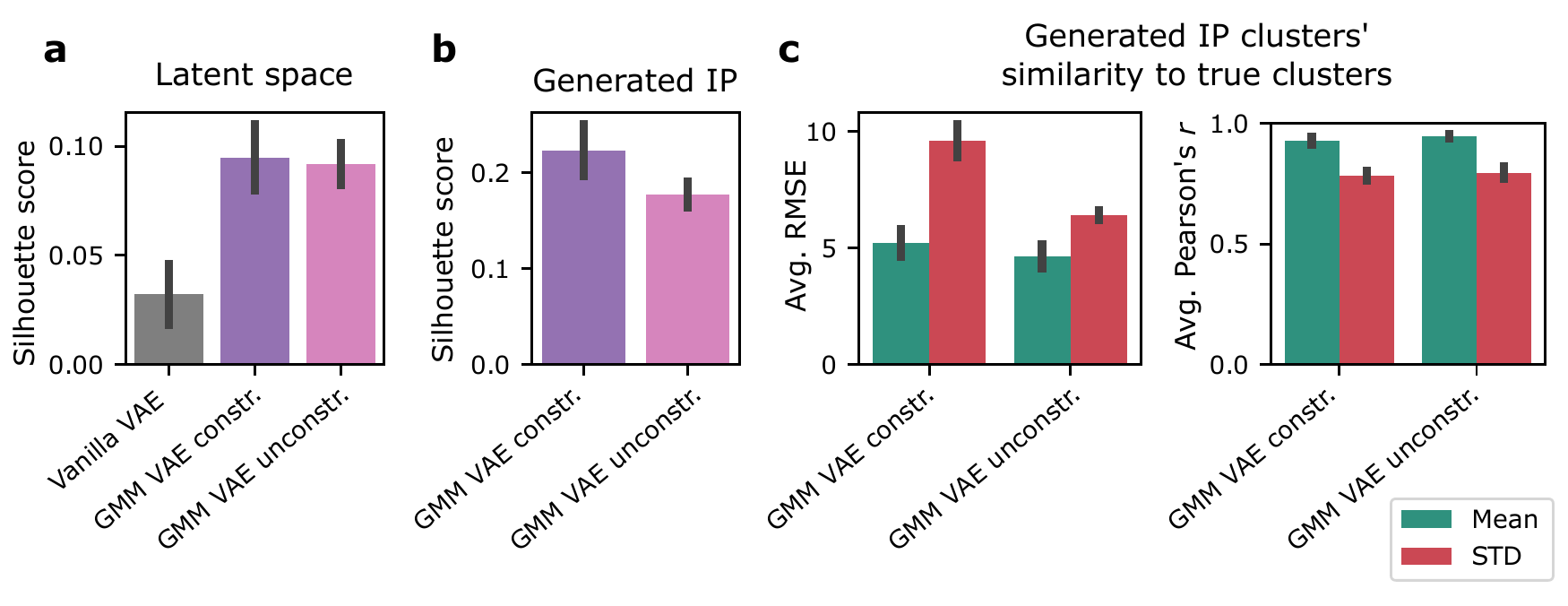}
  \caption{Numerical assessment of models' generative performance. All presented metrics are averaged over five experimental runs, with error bars corresponding to standard deviations across experimental runs. (\textbf{a}) Silhouette score of latent representations, with compounds' guiding labels  as compounds' true clusters. (\textbf{b}) Silhouette scores of generated inhibition profiles, with GMM components from which samples are drawn  as true clusters. (\textbf{c}) Average RMSE (left panel) and Pearson correlation (right panel) between true and generated feature-wise, within-cluster IPs' statistics, shown for cluster means (centroids) and STDs. Average metrics are taken by first averaging over all three clusters and next over the experimental runs.}
  \label{fig:generative_performance_numerical}
\end{figure}

Figure \ref{fig:latent_spaces} compares the structures of the latent spaces of the three considered models. It is clear that in both GMM VAE models the clustering defined by the guiding labels is preserved in the latent space (Fig. \ref{fig:latent_spaces} b, c), i.e. points with the same guiding label are grouped together. By visual assessment, the clusters are the most clearly separated for the GMM VAE unconstrained model (Fig.~\ref{fig:latent_spaces}c). This is also reflected by the corresponding Silhoutte scores that are much higher for GMM models than for Vanilla VAE, with the highest one obtained by GMM VAE constrained (Fig. ~\ref{fig:generative_performance_numerical}a). Interestingly, the latent clustering is to some extent preserved also for the Vanilla VAE model version (Fig. \ref{fig:latent_spaces}a). This suggests that the sole presence of the IP decoder encourages compounds with similar IPs to group together. However, the use of the GMM prior imposes that explicitly. Most importantly, the GMM prior defines the clusters of latent drug representations by associating them to the GMM components, with each guiding label obtaining its separate cluster. In this way, first, we are able to assign a label to a new drug by finding its latent representation and component. Second, we are able to generate new drugs with a pre-specified guiding label.
\begin{wrapfigure}{R}{0.5\textwidth}
  \centering
  \includegraphics[width=0.80\linewidth]{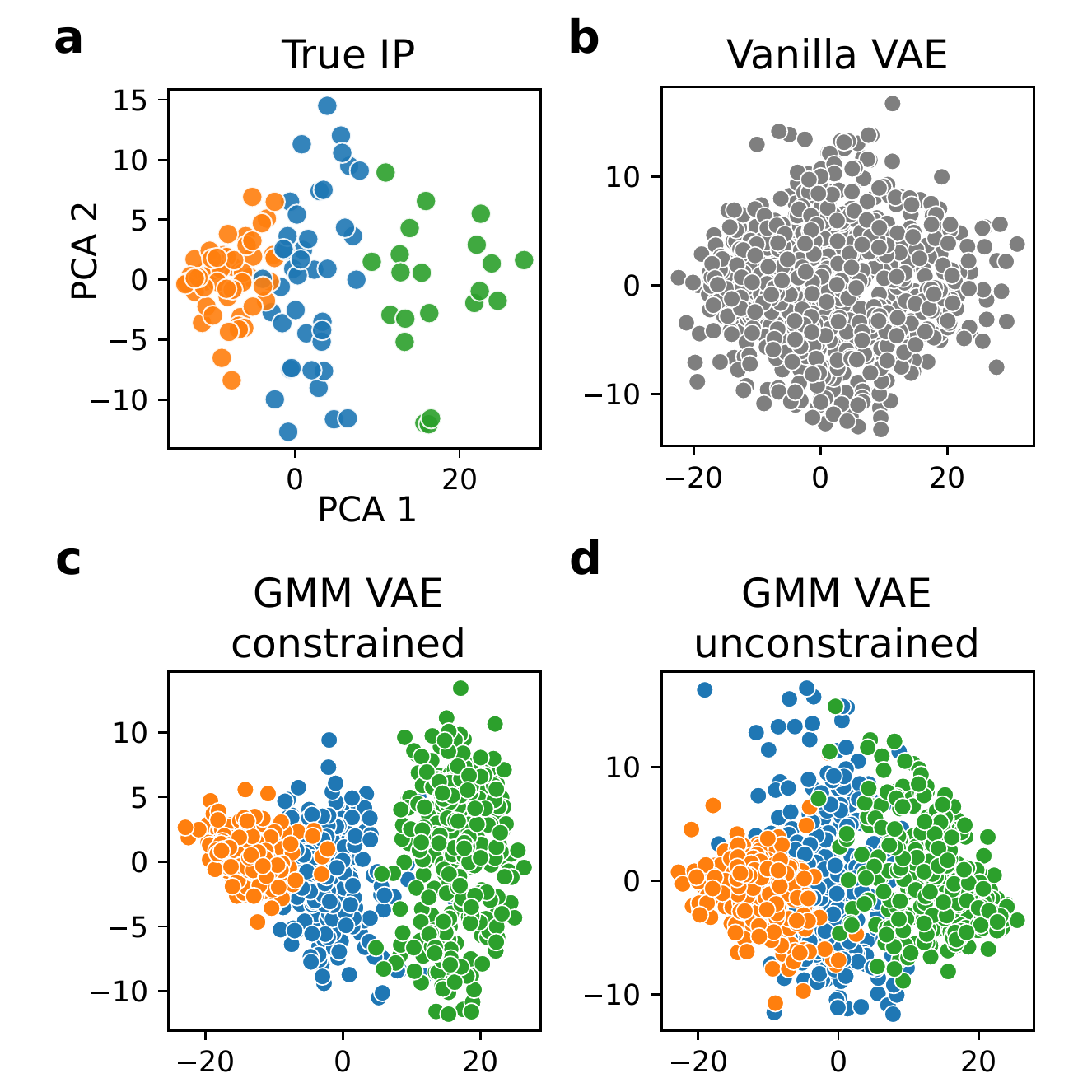}
  \caption{True and generated inhibition profiles visualized in 2D. (\textbf{a}) The true IPs for the 117 available drugs. (\textbf{b}) 900 IPs generated from the Vanilla VAE. (\textbf{c}) IPs generated from the GMM VAE constrained model. 300 samples are drawn. (\textbf{d}) IPs generated from the GMM VAE unconstrained model. Again, 300 samples are drawn per-component. Colors correspond to guiding label or a corresponding GMM component, see Fig. \ref{fig:latent_spaces} for legend.}
  \label{fig:generated_ip}
\end{wrapfigure}

\newpage
\subsection{Generative performance}
In a VAE, new data points can be generated by first sampling from the latent prior and then passing each sample $\vec{z}$ to a decoder. The use of a GMM instead of a normal prior allows to perform this process more precisely; the sample $\vec{z}$  can be obtained from a given, $k$th Gaussian component, which should reflect the actual properties of the compounds in the guiding data space. This is the case in this study, where the guiding labels stem from the clustering in the space of the drugs' IPs. In Fig. \ref{fig:generated_ip}, we verify this hypothesis by visualizing the IPs of the generated samples.

The IPs generated by the Vanilla VAE model do not form any particular clusters (Fig. \ref{fig:generated_ip}b). In contrast, the samples generated from both GMM VAEs clearly confirm that the above assumption is correct; samples generated from different components are also distinguishable after decoding, i.e. the information regarding the latent component is preserved in the true data space (Fig. \ref{fig:generated_ip}c, d). Interestingly, 
not only the grouping of the data points into clusters, but even the actual mutual spatial arrangement of those clusters is preserved between the true IPs (Fig. \ref{fig:generated_ip}a), latent space (Fig. \ref{fig:latent_spaces} b, c), and the generated IPs (Fig. \ref{fig:generated_ip}c, d). However, note that the points are visualized after PCA. Fixing GMM components' $\Sigma_k$ to $I$ impacts the generated IPs; GMM VAE constrained produces more concise and better-separated clusters than unconstrained (Fig. \ref{fig:generated_ip}c, d), which is also reflected by corresponding Silhouette scores (Fig. \ref{fig:generative_performance_numerical}). 

\begin{figure}[h!]
  \centering
  \includegraphics[width=0.88\linewidth]{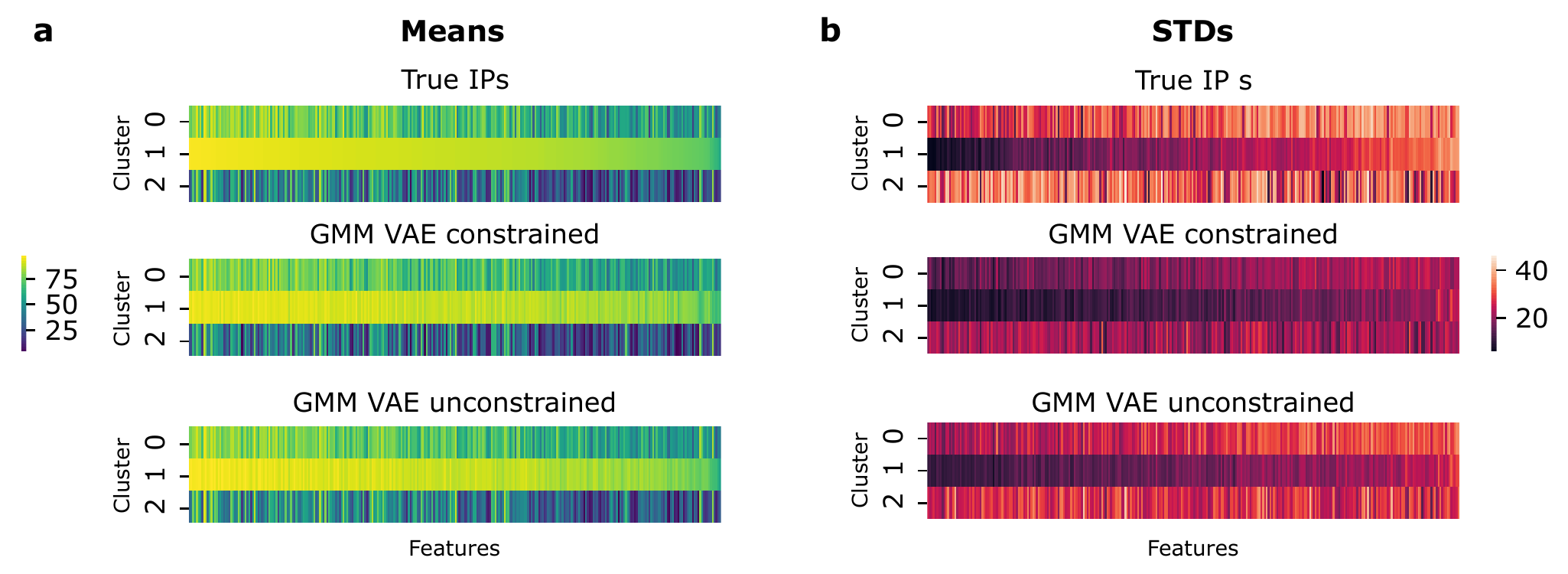}
  \caption{True and generated IPs' feature-wise, within-cluster (\textbf{a}) means and (\textbf{b}) STDs.}
  \label{fig:generated_feat_tables}
\end{figure}

The similarities between the true and generated IPs can also be shown without resolving to dimensionality reduction methods by computing per-cluster, feature-wise statistics such as mean or standard deviation (STD) (Fig. \ref{fig:generated_feat_tables}), where feature-wise means can be thought of as clusters' centroids.  
The IPs generated by both variants of GMM VAE exhibit an excellent concordance with true data in terms of cluster means (Fig. \ref{fig:generated_feat_tables}a). This is apparent on both absolute values level and correlation across features within a given cluster. For example, for a true data, cluster 2 in general corresponds to relatively high inhibition, but for some kinases (features) inhibition is low, and low inhibition of exact same kinases is observed for the generated data.
While differences between GMM VAE constrained and unconstrained in terms of generated IP centroids are hard to assess visually, this is not the case for the within-cluster, feature-wise STDs (Fig. \ref{fig:generated_feat_tables}b). Again, the effect of fixing GMM components' STD is visible; for the constrained model, within-cluster STDs are much smaller compared to the true ones. This also demonstrates that IP decoder has relatively low variance; namely, it is unable to compensate for the low variance of samples from $p(\vec{z})$ in order to bring the generated data's STD closer to the true one. In case of the unconstrained model, the STD is higher and closer to the true one. This clearly demonstrates that in the absence of constraints, the learned components' STDs are larger, and more closely resemble the true data. Indeed, for this particular model, the average value in the covariance matrices ${\Sigma}_k$ is $9.25$. These differences are also clear when assessed by computing the RMSE between true and generated IPs' STD averaged across all three components (Fig. \ref{fig:generative_performance_numerical}c).

In essence, numerical results support all the observations made based on visual assessment. The quality of clustering in the latent space is the highest for GMM VAE constrained model (mean Silhouette score across five experimental runs 0.095), followed by GMM VAE unconstrained and the lowest value obtained by Vanilla VAE (Fig. \ref{fig:generative_performance_numerical}a). Similarly, GMM VAE constrained obtains better clustering quality of generated IPs than the unconstrained version (Silhouette scores 0.223 and 0.177, respectively, Fig. \ref{fig:generative_performance_numerical}b). 

In contrast, GMM VAE constrained obtains slightly worse results than unconstrained in terms of within-cluster statistics similarity between true and generated IPs (\ref{fig:generative_performance_numerical}c). Both constrained and unconstrained models are similarly close to the original data in terms of cluster centroids (average RMSE between true and generated centroids across three clusters of $5.228$ and $4.627$, respectively), especially in terms of correlation (average Pearson correlation $0.928$ and $0.947$, respectively). As in Fig. \ref{fig:generated_feat_tables}b), the differences are more noticeable when considering within-cluster STDs; both models achieve similar correlation ($0.783$ and $0.796$, respectively), but GMM VAE constrained is worse than unconstrained w.r.t RMSE ($9.602$ for constrained and $6.407$ for unconstrained).

\section{Conclusions}   \label{section:conclusions}
In this work, we propose VADEERS, a multi-task generative recommender system for drug sensitivity prediction. The generative part of the model, DVAE, is a variational autoencoder with two decoders and a GMM latent prior. The latent GMM is optimized using guiding labels in order to reflect a given external clustering. This allows to sample data points from a cluster of interest, i.e., having specific desired features. Hence DVAE,  along with other modules, forms a comprehensive model of drugs' and cell lines' properties and interactions, with a guided generative ability. 

{\textbf{Model limitations}} One of the limitations of the proposed model is its inability to generate data points with totally arbitrary features. Namely, the model allows to generate new data points with properties that strictly reflect the clustering observed in the training data. In principle, this could be bypassed by performing various operations on multiple generated data points, however, testing this hypothesis was not in the scope of this analysis. Another important limitation corresponds to the analyzed data; a different choice of data for drugs' representations (e.g. representing SMILES strings as graphs) and guiding data might be more suitable for generating molecule candidates, which, at least in theory, could be synthesized. Both above aspects are directions of future work regarding this study.

\section{Broad impact of the study}  \label{section:broad_impact}
This work introduces several key concepts important for drug sensitivity modeling and compound generation. Still, the proposed model, and more specifically, GMM VAE with semi-supervised clustering with guiding labels, is generic and not limited only to modeling compounds. The notion of optimizing latent space with guiding labels can potentially be beneficial and improve the performance of generative models also in other applications. Moreover, the proposed model offers additional functionality not exploited in this study. For example, setting the number of Gaussian components $K$ greater than number of unique labels $G$ might lead to identification of novel subgroups of samples, not limited to the original choice of guiding labels.

Since VADEERS integrates several sources of data on compounds, cell lines, and drug sensitivity, it can provide important insights regarding these modalities. For example, the usage of the guiding labels and subsequent sampling from a given Gaussian component helps to identify the relationship between variables used as guiding labels and the variables which are decoded by DVAE, i.e, the SMILES embeddings and IPs.
The model comes with the added bonus of the drug sensitivity estimation for these samples for a given cell line or a panel of cell lines. 

In principle, any work on models involving generation of compounds with given properties, including the one presented here, can potentially be used to generate or help to generate harmful chemical agents, such as highly addictive or toxic compounds. In addition, it should be stated that any compound candidate or drug sensitivity estimation originating from an \textit{in silico} model, including the one presented here, is not properly validated in experimental nor clinical setting and should not be directly acted upon without such a validation.

\subsection*{Acknowledgements}
This work was supported by OPUS Grant no. 2019/33/B/NZ2/00956 to ESz from the National Science Centre, Poland,
https://www.ncn.gov.pl/?language=en.

\subsection*{Competing interests}
Merck Healthcare KGaA provides funding for the research group of ESz. ESt is paid by Merck Healthcare KGaA.

\bibliographystyle{unsrt}

\end{document}